\documentclass[10pt,twocolumn,letterpaper]{article}

\usepackage{cvpr}    
\pdfoutput=1

\usepackage{pifont}
\newcommand{\xmark}{\ding{55}} 

\usepackage{amsmath}
\usepackage{array} 

\usepackage[accsupp]{axessibility} 

\usepackage[dvipsnames]{xcolor}

\definecolor{cvprblue}{rgb}{0.21,0.49,0.74}
\usepackage[pagebackref,breaklinks,colorlinks,citecolor=cvprblue]{hyperref}

\usepackage{multicol, multirow}

\def\paperID{14997} 
\def\confName{CVPR}
\def\confYear{2024}

\title{Diff-BGM: A Diffusion Model for Video Background Music Generation}

\author{Sizhe Li$^1$\quad Yiming Qin$^1$\quad Minghang Zheng$^1$\quad Xin Jin$^{2,3}$\quad Yang Liu$^{1}$\thanks{Corresponding author} \\
$^1$Wangxuan Institute of Computer Technology, Peking University \\
$^2$Beijing Electronic Science and Technology Institute \\
$^3$Beijing Institute for General Artificial Intelligence\\
{\tt\small \{lisizhe, minghang, yangliu\}@pku.edu.cn} \quad
{\tt\small kevinqym@stu.pku.edu.cn}  \quad
{\tt\small jinxinbesti@foxmail.com}
}

\begin{document}
\maketitle

\begin{abstract}

When editing a video, a piece of attractive background music is indispensable. However, video background music generation tasks face several challenges, for example, the lack of suitable training datasets, and the difficulties in flexibly controlling the music generation process and sequentially aligning the video and music. In this work, we first propose a high-quality music-video dataset BGM909 with detailed annotation and shot detection to provide multi-modal information about the video and music. We then present evaluation metrics to assess music quality, including music diversity and alignment between music and video with retrieval precision metrics. Finally, we propose the Diff-BGM framework to automatically generate the background music for a given video, which uses different signals to control different aspects of the music during the generation process, i.e., uses dynamic video features to control music rhythm and semantic features to control the melody and atmosphere. We propose to align the video and music sequentially by introducing a segment-aware cross-attention layer. Experiments verify the effectiveness of our proposed method. The code and models are available
at \url{https://github.com/sizhelee/Diff-BGM}.

\end{abstract}

\section{Introduction}
\label{sec:intro}

With the rapid development of multimedia and social platforms, videos become a common way to convey feelings and record lives. When creating videos, to make the video more attractive, a piece of suitable and melodious background music is crucial. However, it is not easy for those who do not have much knowledge of music or video editing to select or create proper or perfectly matched music. What's more, the copyright protection issue has also caused broader public concern. As a result, it is pragmatic to automatically generate background music for a given video. 

\begin{figure}[t]
  \centering
   \includegraphics[width=\linewidth]{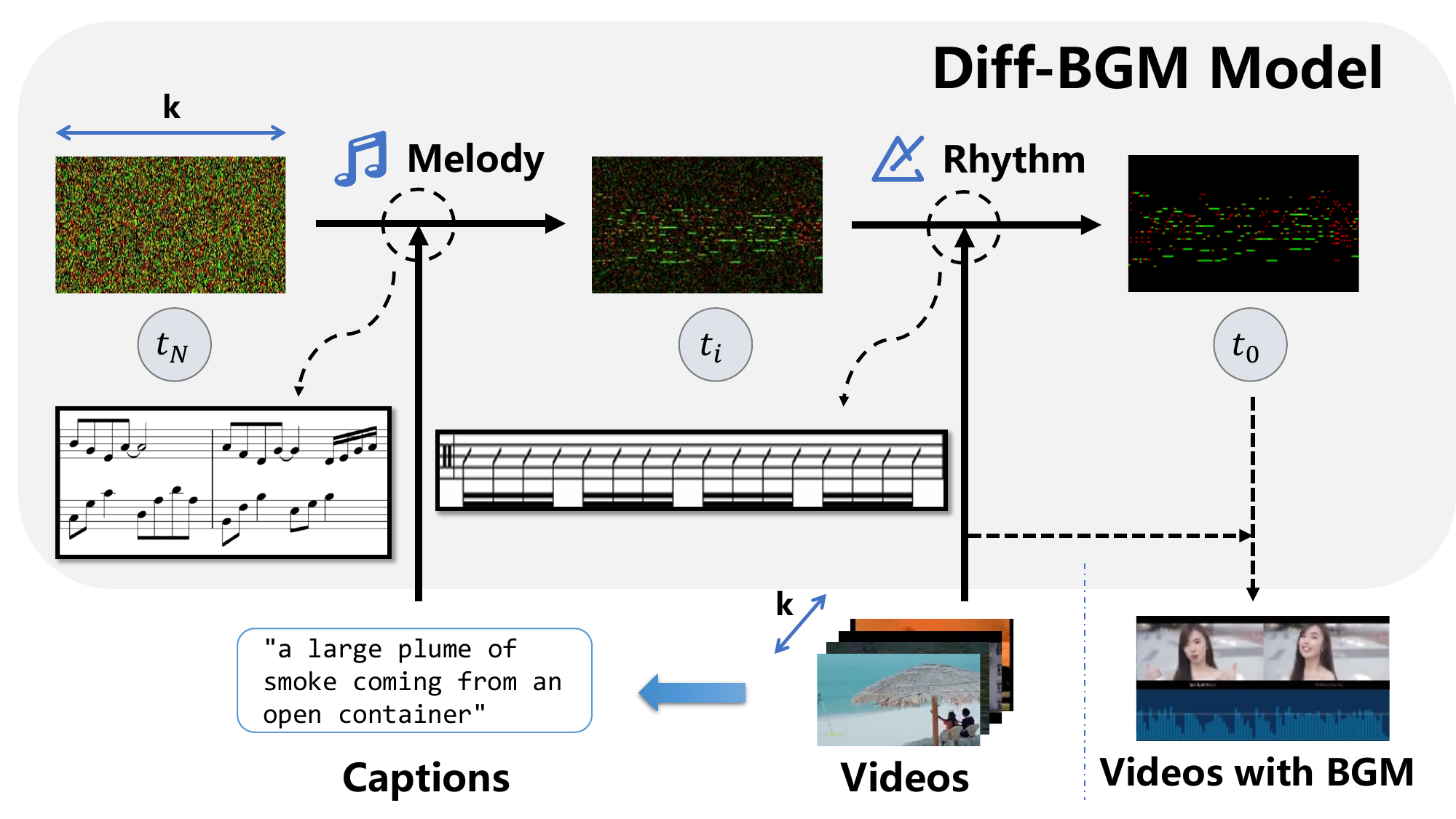}
   \caption{\textbf{Overview of the background music generation process of Diff-BGM.} 
   At different stages of the generation process, Diff-BGM uses different features of videos or captions to control the generation of the music rhythm or melody. We find that the video dynamic feature has more control over the rhythm and the semantic feature has more control over the melody. 
   }
   \label{fig:teaser}
\end{figure}

Existing works~\cite{Wu2021MuseMorphoseFA,Hsiao2021CompoundWT,Dong2022MultitrackMT,Neves2022GeneratingMW,Yu2022MuseformerTW,Agostinelli2023MusicLMGM,Lv2023GETMusicGA,Zhu2022DiscreteCD,Msanii,Moûsai,Mariani2023MultiSourceDM,D3PM,Polyffusion,MusicLDM,Plitsis2023InvestigatingPM} have started focusing on music generation and achieved good results. Besides, some works~\cite{LORIS,D2M-GAN,Gan2020FoleyML} focus on generating music for human-centric videos. In comparison, free-style video background music generation tasks present more challenges. First, to generate proper background music for a video, the model needs to consider multiple aspects of information in the video to \textbf{control different aspects of the music}. In a piece of music, several elements work together to make it pleasant and concordant to listen to. For example, using different rhythms when composing makes the music sound different in dynamic, and different melodies often reflect different atmospheres. Faster music often gives a liveliness or intense feeling, which is suitable for videos that change quickly while slower music tends to be more soothing and is suitable as a warm-style video soundtrack. Sad video clips should correspond to heavy styles and melodies while cheerful videos are matched by upbeat music. \textit{We conclude that visual dynamic changes are linked to music rhythm(the time when the notes appear) and visual semantics influence the melody and atmosphere of the music.}
However, most existing transformer-based models~\cite{CMT,symMV} cannot intuitively reflect the music generation process with corresponding control signals and lack good interpretability. Models for human-centric videos also abstract rhythm information from human motion, which are also unsuitable for freestyle videos.
Secondly, compared to music generation, video-conditioned background music generation requires models to \textbf{temporally align} the video and the music. For example, if a video is composed of many transitions of shot, then each transition should have a more prominent sound to indicate the sudden visual changes.

According to the above challenges, we are the first to consider both fronts. However, there lack suitable datasets. Open-source datasets previously used for other music generation tasks either lack corresponded free-style video samples~\cite{Wang2020POP909AP,Hawthorne2018EnablingFP}, or fail to provide complete annotations of audio or video information~\cite{Hong2017ContentBasedVR,Zhu2022QuantizedGF,Li2021AICM,Li2016CreatingAM,symMV}, making them unsuitable for video background music generation. Details are shown in \cref{tab:dataset_comparison}.
As a result, we collect a video-music dataset named BGM909. Compared to existing datasets, BGM909 has several advantages for background music generation. 
Firstly, we provide high-quality music files, along with comprehensive annotations for various aspects of the audio, such as chords, beats, key signatures, and more. These annotations assist in helping the model learn to analyze music structure and composition. 
Secondly, we offer videos that align with the audio content. Specifically, for song audios, we provide their official MV videos, ensuring semantic consistency between music and video. 
Moreover, our videos undergo manual editing and human checks to ensure perfect temporal alignment with the music.
Additionally, we provide detailed annotations for video including fine-grained natural language descriptions and video shot transitions.
To evaluate the quality of the generated music, we also provide new metrics to measure the music quality and the video-music correspondence.

To tackle the proposed challenges, we propose a framework named \textbf{Diff}usion-based \textbf{B}ack\textbf{G}round \textbf{M}usic generation(\textbf{Diff-BGM}) to generate video-aligned background music. 
For the first challenge, we use diffusion-based models as our framework. It is a recursive process to generate background music so that we can use different signals to control different aspects of music. As shown in \cref{fig:teaser}, to make the music style and atmosphere correspond with the given video, we visualize the music generation process and involve the semantic feature of the video to control the style and melody of the generated music. Then we use the dynamic video feature to control the generation of music rhythm so that the timing information in the two modalities is aligned.
For the second one, we propose a segment-aware cross-attention layer to improve the diffusion framework and sequentially align the video and music. The purpose of temporal alignment is to synchronize the music and video for each segment. Therefore, we introduce cross-attention within the diffusion model and apply time encoding to both modalities. We believe that music generation should be influenced by short-term contexts within the video. Hence, we designed a specific mask to constrain the attention mechanism, obtaining better temporal alignment.

Our contributions are summarized as follows: 
(1) We present BGM909, a high-quality video-music dataset with detailed annotations for background music generation. Also, we provide metrics to measure the video-music correspondence and diversity.
(2) We propose Diff-BGM, the first diffusion-based network for background music generation. It controls the generation process in stages from different dimensions of video and increases the interpretability of the generation process. 
(3) Both objective and subjective evaluation shows that Diff-BGM generates high-quality background music and surpasses the state-of-the-art model.

\section{Related Work}
\label{sec:relatedwork}

\subsection{Music Generation}

In recent years, music generation has attracted much attention, and many models working on music generation have been proposed. 
~\cite{Yu2022MuseformerTW, Neves2022GeneratingMW, Dong2022MultitrackMT, Hsiao2021CompoundWT, Wu2021MuseMorphoseFA} propose to generate music based on transformer and gain satisfying generation results. However, transformer-based models often rely on manually designed tokens to encode music, leading to limited generative ability. With the development of diffusion models, it demonstrates remarkable performance not only in visual tasks but also in music generation. Some works propose diffusion-based model to utilize its excellent generative ability. ~\cite{Msanii,Mariani2023MultiSourceDM,Polyffusion} focus on generating music by exerting control on music while~\cite{Moûsai,MusicLDM,Plitsis2023InvestigatingPM} use text as condition to generate music. \cite{Zhu2022DiscreteCD,Moûsai,MusicLDM,Plitsis2023InvestigatingPM} build bridge between music and other several modals(visual, text, etc). However, those methods do not consider the time alignment between music and the input conditions(like video). As a result, they cannot solve the video background music generation task.

\begin{table*}[t]
    \setlength{\tabcolsep}{4pt}
    \centering
    \renewcommand{\arraystretch}{1.2}
    \centering
    \begin{tabular}{c c c c c c c c c c c}
        \hline
        Dataset & Video & Audio\footnote{Audio records waveform information of sound, while MIDI records music control information on a note-by-note basis.} & MIDI & Style & Chord & Melody & Beat & Caption & Shot dec. & Size \\
        \hline
        MAESTRO~\cite{Hawthorne2018EnablingFP} & \xmark & \checkmark & \checkmark & \xmark & \xmark & \xmark & \xmark & --- & --- & 1,276 \\
        POP909~\cite{Wang2020POP909AP} & \xmark & \xmark & \checkmark & \checkmark & \checkmark & \checkmark & \checkmark & --- & --- & 909 \\
        \hline
        HIMV-200k~\cite{Hong2017ContentBasedVR} & \checkmark & \checkmark & \xmark & \xmark & \xmark & \xmark & \xmark & \xmark & \xmark & 200,500 \\
        TikTok~\cite{Zhu2022QuantizedGF} & \checkmark & \checkmark & \xmark & \xmark & \xmark & \xmark & \xmark & \xmark & \xmark & 445 \\ 
        AIST++~\cite{Li2021AICM} & \checkmark & \checkmark & \xmark & \checkmark & \xmark & \xmark & \xmark & \xmark & \xmark & 1,408 \\
        URMP~\cite{Li2016CreatingAM} & \checkmark & \checkmark & \checkmark & \xmark & \xmark & \xmark & \xmark & \xmark & \xmark & 44 \\
        SymMV~\cite{symMV} & \checkmark & \checkmark & \checkmark & \checkmark & \checkmark & \checkmark & \xmark & \xmark & \xmark & 1,140 \\
        \hline 
        BGM909(Ours) & \checkmark & \checkmark & \checkmark & \checkmark & \checkmark & \checkmark & \checkmark & \checkmark & \checkmark & 909 \\
        \hline
    \end{tabular}
    \caption{\textbf{Comparison between different music datasets.} Our proposed BGM909 contains a considerable amount of video-music pairs, being able to be applied to the background music generation task. Different from existing datasets, BGM909 provides various musical annotations, metadata, captions and shot detection. Two popular music datasets are shown in the first two rows for reference.}
    \label{tab:dataset_comparison}
\end{table*}

\subsection{Background Music Generation}

Some methods~\cite{LORIS,D2M-GAN,Gan2020FoleyML} focus on generating music for human-centric videos (i.e. dance or sports videos), in which the rhythm is largely dependent on human motions and is not accessible in freestyle videos.
The task of video background music generation was first proposed by CMT~\cite{CMT} and has gained more and more attention. Existing background music generation methods mostly use the transformer-based framework and establish the relationship between video and music.
CMT~\cite{CMT} first encodes the chords and notes to give the representation of a piece of music and train the model to understand the logic of music, then it establishes three rhythmic relations ($e.g.$ motion speed of the video corresponds to note density in the audio) between the video and background music to narrow the gap between the two forms of expression. Other than only using the rule-based rhythmic relationships, V-MusProd~\cite{symMV} and Video2Music~\cite{Video2Music} focus on semantic-level correspondence. They extract the semantic feature of the video to control the style of the generated music in the multi-modal transformer blocks.
However, transformer-based methods suffer from the same challenges that it is hard to control the process of end-to-end generation thus leading to poor interpretability. 
We propose a diffusion-based framework to generate background music for videos to make full use of the generative ability of diffusion models. Besides, we use different features to control different aspects during the generation process and conduct temporal alignment between video and music.

\section{Dataset}

Due to the lack of high-quality open-source datasets for background music generation, we collect a new video-music dataset BGM909 based on POP909~\cite{Wang2020POP909AP} containing 909 pieces of piano version music and corresponding well-aligned videos.  
BGM909 has the following advantages over previous datasets.
(1) We provide high-quality MIDI files of music, and detailed annotations such as chords, styles etc. 
(2) The content of the videos aligns with the music. Specifically, we provide the official MV videos for each song music to ensure semantic coherence. (3) We also manually edit and check the video-music pairs to ensure perfect temporal alignment. (4) Detailed annotations for videos including fine-grained language descriptions and shot transitions are provided for further study.
\cref{tab:dataset_comparison} shows the comparison of BGM909 with other existing video-music datasets.

\subsection{Data Collection}

Tons of music-video pairs are available on the Internet. However, it is hard to gain high-quality noiseless audio from those videos. As a result, we start with the existing well-annotated music in POP909~\cite{Wang2020POP909AP} dataset.
For each MIDI file in POP909, we collect its metadata and use the song title and singer as keywords to search for the corresponding official video on YouTube. After downloading the videos, we remove those that only had static interfaces or lyrics. Then for those left videos, we manually edit the video to align it with the MIDI file temporally ($e.g.$ some audios are not played at the beginning of the videos). 
To ensure dataset quality, we manually check the collected video-music pairs.

\subsection{Data Annotation}

\textbf{Melody, Bridge and Piano.} Each MIDI file contains three tracks of melody, bridge, and piano, representing the lead melody transcription, the secondary melody and the main body of the accompaniment separately. 

\textbf{Chord, Beat and Key Signature.} A chord is a number of notes played at the same time, specific notes comprise harmony chords and play an important role in setting the base tone of the music. Beats mean the length of each note and are basic components of music rhythm. Key signature represents the tonality of the music. We provide chord, beat, and key signatures separately, including beat and downbeat annotations, start and end time for each chord, and chord names. Detailed algorithms can be found in~\cite{Wang2020POP909AP}. 

\textbf{Natural Language Descriptions.} We provide fine-grained natural language descriptions for each video in BGM909. We generate 10 description sentences for each 8 frames in each video with a pre-trained BLIP~\cite{Li2022BLIPBL} model. 
It also serves as a lightweight substitute for video features, capable of fully expressing the semantic information of the video. Besides, the captions can be extended to other tasks like text-to-music generation.

\textbf{Camera Shot Detection.} We extract each shot of the video based on the camera switching. Music often stands out at the transitions in the video, therefore, captured shot transitions often have a significant impact on the rhythm and variations of the music. The timing of shot transitions is also a focal point to pay attention to during music generation. Shot detection assists in training the alignment between music and video. Videos have 62 shots on average.

\textbf{Styles.} We provide GPT with the song's name and the associated artist and obtain the style classification of each audio to further improve the dataset and prove the generality of BGM909. The songs are from 646 different artists and are divided into 8 different styles in total.

\textbf{Metadata.} We also provide extra metadata for the music in BGM909 compared with POP909, like lyrics, genre, and rhythmic pattern. The metadata information is useful for data analysis and may be used in future research works like music-to-video generation.

\begin{figure*}[t]
  \centering
   \includegraphics[width=0.99\linewidth]{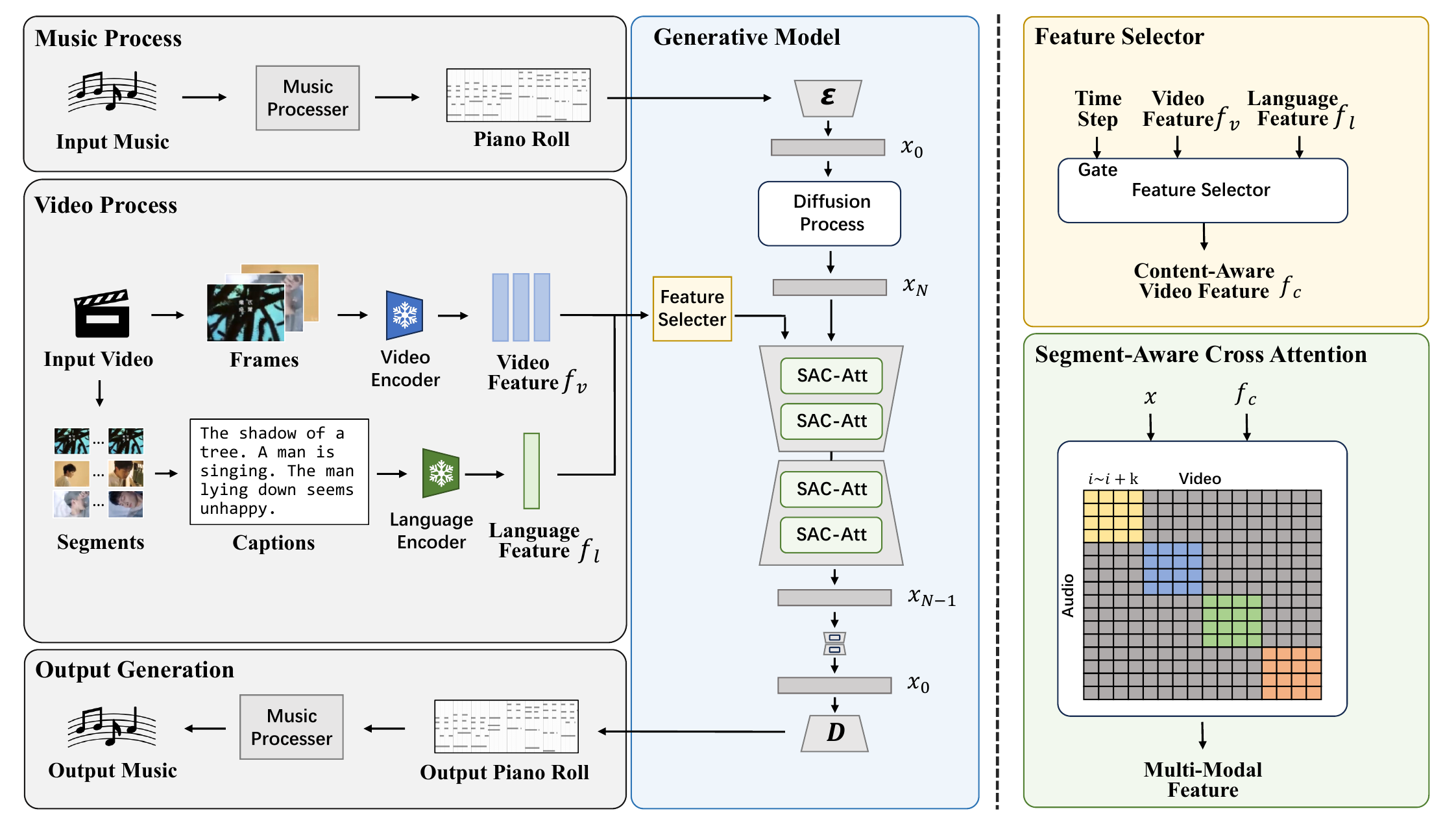}

   \caption{\textbf{Illustration of our Diff-BGM model.} We process the input music and video, gain piano rolls to represent the music and extract visual features to represent the videos. In order to get richer semantic information, we segment the video and generate captions then extract language features. The backbone of the generation model is a diffusion model, with the processed visual and language features as conditions to guide the generation process. We propose a feature selector to choose features to control the generation process. And to better align the timing of the music and video, we design segment-aware cross-attention layer to grasp the timing feature in different modalities.}
   \label{fig:method}
\end{figure*}

\section{Method}

We propose a novel music generation pipeline named Diff-BGM following the principles of latent diffusion models~\cite{latentdiffusionmodel} to deal with the video background music generation task. The framework of Diff-BGM is shown in~\cref{fig:method}(left), which contains a Music Process Module, a Video Process Module, a Generative Model and an Output Generation Module. The music process module takes original midi files as input and generates corresponding piano rolls to represent the music. The video process module takes the original videos as input. To guide the generation process, we extract the visual features of the video, generate captions for the video and extract language features for those captions. The generative module is a diffusion model, which uses the extracted features as conditions to generate a new piano roll. In the end, after gaining the generated piano roll, the output generation module is used to process the piano roll and generate the corresponding music.
In order to control different stages of music generation using different features, we introduced a feature selector, as shown in \cref{fig:method}(right).
To consider the timing in the video and the music, and align the video and the generated music, we introduce segment-aware cross-attention to align the video and music. 

\subsection{Polyffusion Baseline Revisited}
\label{subsec:ddpm}
We use Polyffusion~\cite{Polyffusion} as our baseline. It is trained on midi data and outputs a midi file for unconditional generating. Polyffusion is a diffusion-based music generation framework, with a Denoising Diffusion Probabilistic Model~\cite{ddpm} as its generative baseline and provides a complete midi process algorithm. We follow the structure and data process algorithm proposed by Polyffusion.

\textbf{Music Process.} 
Polyffusion divides the input midi music into 8-bar (32-beat, $T_0=128$ time steps) segments and transfers it into image-like piano roll representation $x\in\mathbb{R}^{2\times T_0\times P}$, which is a 2-channel binary tensor. 
The MIDI pitch ranges 0...127 so we gain $P=128$ pitch bins. 
In the piano roll representation, entry $a(c, t, p)$ represents at time step $t$ and MIDI pitch $p$ whether there is a note onset($c=0$) or sustain($c=1$).

\textbf{Generative model.}
Polyffusion uses a latent diffusion model~\cite{latentdiffusionmodel} as generator to generate piano rolls for videos. 
It contains a diffusion process and a denoising process. In the diffusion process, the structure of data $x_0$ is broken up step by step by iteratively adding Gaussian noises in $N$ steps:
\begin{equation}
    q(x_t|x_{t-1})=\mathcal{N}(x_t;\sqrt{1-\beta_t}x_{t-1},\beta_t\mathbf{I})
\end{equation}
\begin{equation}
    q(x_{1:N}|x_0)=\prod \limits_{t=1}^N q(x_t|x_{t-1})
\end{equation}
where $\beta_1, \beta_2, ..., \beta_N$ are a set of variance scheduling parameters, $x_0$ is the clean input piano roll.
Then in the denoising process, the model learns to reconstruct the original structure of $x_0$ from the noisy input $x_N\sim \mathcal{N}(0, \mathbf{I})$. It is defined as a Markov chain with learned Gaussian transitions:
\begin{equation}
    p_\theta(x_{t-1}|x_t)=\mathcal{N}(x_{t-1};\mu_\theta(x_t, t), \sigma_\theta(x_t, t))
\end{equation}
\begin{equation}
    p_\theta(x_{0:N})=p(x_N)\prod\limits_{t=1}^N p_\theta(x_{t-1}|x_t)
\end{equation}
The training process is performed  by optimizing the following target:
\begin{equation}
    L(\theta)=\mathbb{E}_{x_0,\epsilon,t}[||\epsilon-\epsilon_\theta(\sqrt{\Bar{\alpha}_t}x_0+\sqrt{1-\Bar{\alpha}_t}\epsilon, t)||^2]
\end{equation}
where $\epsilon_\theta$ represents the model parameters, $t$ is uniform between $1$ and $N$, $\epsilon\sim\mathcal{N}(0,I)$, $\alpha_t=1-\beta_t$, $\Bar{\alpha}_t=\prod_{i=1}^t\alpha_i$.

Although Polyffusion has the ability to generate music unconditionally, it still cannot generate background music for a given video and has not addressed the two challenges mentioned in \cref{sec:intro}, namely the inability to achieve conditional control and alignment between music and video. Therefore, to achieve temporal alignment between music and video, we made improvements upon Polyffusion.

\subsection{Video Process}

Diff-BGM model takes videos $V$ as input. We first sample $T$ frames and use a pre-trained video encoder to extract the visual feature $F_v\in\mathbb{R}^{T\times d_1}$ of each frame. Besides, we segment the video into $T$ segments $V=\{S_1, S_2, ..., S_T\}$ and generate natural language captions $C_i$ according to the video content for each segment $S_i$. For the captions, we use a pre-trained language encoder to extract the language features $F_l\in\mathbb{R}^{T\times d_2}$ for the captions.

\subsection{Feature Selection.} 

We analyze the timestep intervals defined as~\cite{Patashnik2023LocalizingOS} to show the type of attribute in the music controlled by each interval. 
By observing the unconditional music generation process, we found that models tend to generate the melody, which is influenced by the semantics, and then generate the rhythm of the music, which is related to the dynamic feature of the video.
As a result, at different timestep intervals, we use different features as conditions to describe the video. 

The final condition feature $F_c$ is represented as:
\begin{equation}
    F_c = \left\{\begin{array}{ll}
        F_l, & \text{TimeStep} > t_0 \\
        F_v, & \text{TimeStep} \leq t_0
    \end{array} \right.
\end{equation}
where $t_0$ is a hyper-parameter representing the key time step during the denoising process decreasing from $N$ to 0, Timestep represents the current step of denoising, $F_c$ is the condition feature of the video. 

\subsection{Sequential Attention}
\label{subsec:ldm}
We aim to generate music for given videos, which means the style and atmosphere of the music should match the semantic content of the video. Besides, when shot changes or noticeable motion changes exist, there should be a homologous response in the music. Obviously, unconditional diffusion and denoising process cannot achieve this goal.
Firstly, as we require the generated music to match the given video in terms of rhythm, melody, and other aspects, we introduce video features as conditions. We also propose a segment-aware cross-attention layer to fuse the video features with music features in the latent space of the diffusion model, ensuring that the generation process is consistently guided by the video, resulting in music related to the video. 
Additionally, to achieve fine-grained temporal alignment between music rhythm and the video, we applied time encoding to both and introduced specially designed masks to conduct sequential attention. These enable the music generation process to incorporate small-scale video features as context, facilitating precise alignment between the two and generating high-quality background music.

In order to align video and music sequences and understand the context information in both modalities, we follow Latent Diffusion~\cite{latentdiffusionmodel} and design a segment-aware cross-attention layer. The input noisy latent representation $x_t$ serves as Query, while the condition feature $F_c$ serves as Key and Value. Then the attention can be represented as:
\begin{equation}
    Attn = \frac{Q K^\top}{\sqrt{d_{key}}}
\end{equation}
where $Q, K$ denote Query and Key separately, $d_{key}$ is the dimension of the condition feature.
However, to align the two modalities in timing, long-term context is not so important as short-term context for music is often associated with the current clip of the video. As a result, a special mask is designed to only pay attention to short-term context and neglect long-term context.
As shown in~\cref{fig:method}, we divide adjacent $k$ frames into short-term contexts, then only the features of those $k$ adjacent frames can influence the generated music at each time spot. The mask is given as follows:
\begin{equation}
    \mathcal{M}ask_{i,j}=\left\{\begin{array}{ll}
        1, & k\cdot\gamma\leq i,j<k\cdot(\gamma+1)  \\
        0, & \text{otherwise}
    \end{array} \right.
\end{equation}
where $\mathcal{M}ask\in\mathbb{R}^{T\times T}$ represents the designed attention mask, $\gamma\in\{0,1,..,\frac{T}{k}\}$ represents the number of contexts.

As a result, the output of the segment-aware cross-attention layer is as follows:
\begin{equation}
    x_{out} = \underset{seq}{softmax}\Bigg(\mathcal{M}ask\Bigg(\frac{Q K^\top}{\sqrt{d_{key}}}\Bigg)\Bigg)V
\end{equation}
where $V$ denotes Value. Then the output $x_{out}$ combines the short-term context features of both video and music so that the model is able to generate video-aligned music.

\begin{table*}[t]
\centering
\begin{tabular}{cccccccc}
\hline
\multicolumn{1}{c}{\multirow{2}{*}{Methods}} & \multicolumn{3}{c}{Music Quality} & \multicolumn{3}{c}{Video-Music Correspondence} & \multicolumn{1}{c}{\multirow{2}{*}{Diversity$\uparrow$}} \\
\multicolumn{1}{c}{} & PCHE$\rightarrow$ & GPS$\rightarrow$ & SI$\rightarrow$ & P@5$\uparrow$ & P@10$\uparrow$ & P@20$\uparrow$ & \multicolumn{1}{c}{} \\ 
\hline
Real(BGM909) & 2.717 & 0.708 & 0.486 & --- & --- & --- & 6.664\\ 
\hline
CMT~\cite{CMT} & 2.398 & 0.594 & 0.761 & 10.56\% & 18.59\% & 36.40\% & 6.025\\ 
Riffusion~\cite{Riffusion} & 2.556 & 0.509 & 0.412 & 9.12\% & 15.71\% & 34.29\% & 6.335 \\
\hline
\multicolumn{1}{l}{Unconditional} & 3.189 & 0.595 & 0.528 & 8.08\% & 15.89\% & 31.93\% & \textbf{6.421}\\

\multicolumn{1}{r}{+ Video feature} & \underline{2.835} & 0.514 & 0.396 & \textbf{13.44\%} & \underline{23.54\%} & \underline{43.20\%} & 5.742\\

\multicolumn{1}{r}{+ Feature selection} & \textbf{2.721} & \textbf{0.789} & \underline{0.523} & 11.00\% & 20.79\% & 38.47\% & 5.246\\

\hline
\multicolumn{1}{r}{+ SAC-Attn (Diff-BGM)} & 2.840 & \underline{0.601} & \textbf{0.521} & \underline{13.28\%} & \textbf{23.91\%} & \textbf{44.10\%} & 5.153\\
\hline
\end{tabular}
\caption{\textbf{Objective evaluation results on BGM909 test set.} We evaluate both music quality and video-music correspondence with several metrics, where P indicates precision, the higher P is better. PCHE indicates Pitch Class Histogram Entropy, GPS indicates Grooving Pattern Similarity and SI means Structureness Indicator, where closer to Real is better.}
\label{tab:obj}
\end{table*}

\begin{table*}[t]
\centering
\begin{tabular}{ccccccc}
\hline
\multicolumn{1}{c}{\multirow{2}{*}{Methods}} & \multicolumn{2}{c}{Music Quality} & \multicolumn{4}{c}{Video-Music Correspondence} \\
\multicolumn{1}{c}{} & PCHE$\rightarrow$ & SC$\rightarrow$ & P@5$\uparrow$ & P@10$\uparrow$ & P@20$\uparrow$ & AR$\downarrow$\\ 
\hline
Real(official test set) & 2.633 & 0.986 & --- & --- & --- & ---\\ 
CMT~\cite{CMT} & 2.444 & 0.990 & 8.9 & 17.7 & 31.0 & 33.4\\
V-MusProd~\cite{symMV} & 2.607 & 0.983 & \underline{15.7} & \underline{24.6} & \underline{44.8} & \underline{25.4}\\ 
\hline
Real(our test set) & 2.538 & 0.882 & --- & --- & --- & ---\\
Diff-BGM & 2.738 & 0.878 & \textbf{19.0} & \textbf{28.6} & \textbf{47.6} & \textbf{19.4}\\
\hline
\end{tabular}
\caption{Objective Evaluation on SymMV test set. We evaluate music quality with Pitch Class Entropy and Scale Consistency and evaluate video-music correspondence, where P represents precision, AR represents average
rank of the ground truth video}.
\label{tab:symmv}
\end{table*}

\subsection{Train and Inference}

Following the strategy above, the final training objective is
\begin{equation}
    L_{cond}(\theta)=\mathbb{E}_{x_0, F_c, \epsilon, t}[||\epsilon-\epsilon_\theta(\sqrt{\Bar{\alpha}_t}x_0 + \sqrt{1-\Bar{\alpha}_t}\epsilon, t, F_c)||^2].
\end{equation}
where $x_0$ represents the original clean piano roll, $F_c$ denotes the condition feature, $t$ denotes the time step.
During inference, Diff-BGM receives a random noise as input $x_N$ and uses the video dynamic and semantic features as conditions. We can control the generation process by flexibly adjusting the key time step $t_0$ to select the condition features and generate diverse music for a video.

\section{Experiments}

\subsection{Implementation Details}

To make a fair comparison, we follow previous work~\cite{ddpm,Polyffusion} to use a Gaussian noise schedule and the noise prediction objective in \cref{subsec:ddpm} for all experiments. Our segment-aware cross-attention layers are set as~\cite{latentdiffusionmodel}. The diffusion step $N$ is set to 1,000. Diff-BGM model converges around 100 epochs on Adam Optimizer~\cite{Kingma2014AdamAM} with a constant learning rate 5e-5. We use official pre-trained Video CLIP~\cite{Xu2021VideoCLIPCP} as video encoder to extract visual features. We choose the BLIP~\cite{Li2022BLIPBL} model to segment the video and use official pre-trained bert-base-uncased model~\cite{BERTbaseuncased} as the language encoder. The visual and language encoders keep frozen during training. In the segment-aware cross-attention module, we set $k$ to 8 and $t_0$ to 200.

\begin{table}[t]
\centering
\begin{tabular}{cc|cc}
\hline
\multicolumn{2}{c|}{\multirow{2}{*}{Metrics}} & \multicolumn{2}{c}{Rates}\\
\multicolumn{2}{c|}{} & Experts & Non-Exp.\\ 
\hline
\multicolumn{1}{c|}{\multirow{2}{*}{Music Quality}} & M. Melody & 75.0\% & 81.5\%\\
\multicolumn{1}{c|}{} & M. Rhythm & 63.9\% & 74.4\% \\ 
\hline
\multicolumn{1}{c|}{\multirow{2}{*}{\begin{tabular}[c]{@{}c@{}}Video-Music \\ Correspondence\end{tabular}}} & V. Content & 75.0\% & 80.4\%\\
\multicolumn{1}{c|}{} & V. Rhythm & 70.4\% & 75.0\%\\ 
\hline
\multicolumn{1}{c|}{\multirow{2}{*}{Expertise}} & Chord & 70.4\% & ---\\
\multicolumn{1}{c|}{} & Accom. & 78.7\% & ---\\ 
\hline
\multicolumn{1}{c|}{Overall} & Overall Pref. & 77.8\% & 83.9\%\\ 
\hline
\end{tabular}
\caption{\textbf{Subjective evaluation.} The preference rates for Diff-BGM against CMT~\cite{CMT} are shown in music quality metrics, video-music correspondence metrics, and expertise metrics.}
\label{tab:sub}
\end{table}

\begin{table*}
\centering
\begin{tabular}{c|cccccccc}
\hline
              & M. Mel & M. Rhy & V. Content & V. Rhy & Chord & Accom. & Overall & Conf.\\ 
\hline
Human   & 3.48\scriptsize$\pm$0.99 & 3.46\scriptsize$\pm$0.99 & 3.45\scriptsize$\pm$1.25 & 3.28\scriptsize$\pm$1.06 & 3.26\scriptsize$\pm$0.97 & 3.39\scriptsize$\pm$1.09 & 3.48\scriptsize$\pm$1.02 & 3.90\\
Riffusion\cite{Riffusion} & 2.97\scriptsize$\pm$1.11 & 2.86\scriptsize$\pm$1.02 & 2.74\scriptsize$\pm$1.30 & 2.68\scriptsize$\pm$1.15 & 2.89\scriptsize$\pm$1.02 & 2.86\scriptsize$\pm$1.08 & 2.94\scriptsize$\pm$1.13 & 3.88\\
CMT~\cite{CMT}     & 2.94\scriptsize$\pm$1.04 & 2.97\scriptsize$\pm$1.13 & 2.74\scriptsize$\pm$1.22 & 2.72\scriptsize$\pm$1.19 & 2.81\scriptsize$\pm$1.02 & 2.78\scriptsize$\pm$1.12 & 2.88\scriptsize$\pm$1.10 & 3.73\\
ours    & \textbf{3.29}\scriptsize$\pm$0.87 & \textbf{3.22}\scriptsize$\pm$0.96 & \textbf{3.14}\scriptsize$\pm$1.03 & \textbf{3.16}\scriptsize$\pm$1.09 & \textbf{3.11}\scriptsize$\pm$1.03 & \textbf{3.22}\scriptsize$\pm$0.97 & \textbf{3.33}\scriptsize$\pm$0.99 & \textbf{3.78}\\

\hline
\end{tabular}
\caption{\textbf{Subjective evaluation.} Experts are asked to score 6 music generated by different models and also created by humans.}
\label{tab:sub_new}
\end{table*}

\subsection{Objective Evaluation}

\textbf{Metrics.}
As for evaluating metrics for the task of generating background music for videos, they have not been fully refined to date. Therefore, building upon existing metrics, we have proposed additional metrics to assess the generated results of background music as follows:

\begin{itemize}
    \item \textbf{Music Quality.} We choose the same metrics as~\cite{Wu2020TheJT,CMT} to evaluate music quality, including \textit{Pitch Class Histogram Entropy(PCHE)} which measures the uncertainty of the distribution of the notes and reflects the quality of tonality, \textit{Grooving Pattern Similarity(GPS)} which measures the quality of the rhythmicity, and \textit{Structureness Indicator(SI)} which captures the repetition in the music by measuring the overall structure and reflects the catchiness and the emotion-provoking nature~\cite{BrainonMusic}. On SymMV~\cite{symMV} dataset, scale consistency(SC) is also used to evaluate the music quality. Note that the overall quality is not indicated by how high or low these metrics are, but instead by their \textit{closeness} to the real music data. 

    \item  \textbf{Diversity} We propose a metric to evaluate the diversity of the generated music. We randomly divide the generated music into two subsets, $S_d$ samples in each set. The diversity of the generated music is defined as:
    \begin{equation}
        \text{Diversity} = \frac{1}{S_d}\sum\limits_{i=1}^{S_d}||v_i-v_i'||_2
    \end{equation}\
    where $v_i, v_i'$ represent the music feature of the $i-$th sample in the two subsets separately.

    \item \textbf{Music Retrieval} We propose a new metric to measure the music-video consistency. Given a piece of generated music $\hat{m}$ and the ground truth music of its condition video $m$, we randomly select $M-1$ pieces of music $m_i$. Here we use Musicnn~\cite{Pons2019musicnnPC} to extract music feature for each generated item. If the ground-truth music ranks in the top-$K$ place, then we consider it a successful retrieval. All generated samples are used to calculate the successful retrieval rate as the final precision score $P@K$. Here we set $M=64, K=5, 10, 20$. Since the ground truth music is related to the given video, the proposed retrieval precision metric is able to measure how well the generated music aligns with the given video.
\end{itemize}

\textbf{Results.} The results on BGM909 test set are shown in \cref{tab:obj}. Compared with CMT~\cite{CMT}, our Diff-BGM surpasses it on both music quality metrics and video-music correspondence, and has a gain of $4.91\%, 8.15\%, 10.39\%$ on the retrieval metircs, which proves that Diff-BGM generates higher-quality music and has a better understanding of the correspondence between video and music.

Results on SymMV~\cite{symMV} dataset are shown in \cref{tab:symmv}. For comparison, we have re-divided the train/val/test sets based on the instructions provided in~\cite{symMV}, ensuring that the size of each set aligns with the proposed official sizes\footnote{As of now, SymMV has not publicly disclosed complete information, including the partitioning of train/val/test sets and the alignment timestamps between video and audio.}.
Methods in the first block in \cref{tab:symmv} are evaluated on the official SymMV test set, while methods in the second block are on the test split we obtained \footnote{We are unable to present the performance results of V-MusProd on the new split since their code is not publicly accessible.}. Since the test split is not identical \footnote{The numbers of real data on the two splits are different}, direct numerical comparisons of music quality metrics cannot be made. However, it's important to note that retrieval-based metrics (for video-music correspondence evaluation) can still be compared in a relatively fair manner, given that the size of the retrieval pool is consistent. As shown in \cref{tab:symmv}, Diff-BGM outperforms CMT and V-MusProd in video-music correspondence metrics by a large margin, indicating that Diff-BGM can effectively align video and music during the generation process.

\subsection{Subjective Evaluation}

The best way to evaluate a generative model today remains using user study, and it is widely adopted in previous works~\cite{CMT,symMV,Ruan2022MMDiffusionLM,Wu2020TheJT}. We conduct the user study by designing and sending out questionnaires. We invite 46 people to participate in the user study, 18 of them are experts with expert knowledge in music, 28 are non-experts. We choose videos from different categories then use Diff-BGM and CMT to generate music for each video separately, present them randomly for blindness, and require the participants to compare the two generation results in several aspects and give preference scores separately. For some videos are long, the questionnaire takes about 25 minutes to complete.

\textbf{Metrics.} For each video, participants are required to listen to several music pieces and score them from several aspects as~\cite{symMV}: (1)Music Melody: the richness of the musical melody; (2)Music Rhythm: the structure consistency of rhythm; (3) Content Correspondence: the correspondence between music and video content; (4) Rhythm Correspondence: the correspondence between music and video rhythm; (5) Overall Preference. Besides, the experts are asked to evaluate two extra metrics related to music theory: (6) Chord Quality: the quality, composition and degree of harmony of generated chords; (7) Accompaniment Quality: the richness and quality of the generated accompaniment. 

\textbf{Results.} The result is provided in \cref{tab:sub}, showing the preference rate of Diff-BGM against CMT (the percentage of participants who consider music generated by Diff-BGM better than CMT). It shows that in all metrics and user groups, Diff-BGM outperforms CMT($>$50\%), indicating that Diff-BGM generates higher-quality music and better understands video-music correspondence. We also include preference scores of more models to compare as shown in \cref{tab:sub_new}. It can be observed that in every aspect, Diff-BGM is the closest to artificial(line 1). Note the \textit{Human}-created music score is only 3.5. It reflects how much improvement is needed to get human-level creation. 

\begin{table}[t]
\centering
\begin{tabular}{cccc}
\hline
\multicolumn{1}{c}{\multirow{2}{*}{Methods}} & \multicolumn{3}{c}{Music Quality}\\
\multicolumn{1}{c}{} & PCHE$\rightarrow$ & GPS$\rightarrow$ & SI$\rightarrow$ \\ 
\hline
Only Video & 2.835 & 0.514 & 0.396\\
Only Language & 2.849 & \textbf{0.641} & 0.521 \\
Video+Language & 2.840 & 0.626 & 0.536\\
Language+Video & \textbf{2.781} & 0.601 & \textbf{0.517}\\
\hline
\end{tabular}
\caption{\textbf{Ablation studies on feature selector.} We use different features in lines 1-2 and different feature orders in lines 3-4 to control the generation process. Closer to Real is better.}
\label{tab:ablation}
\end{table}

\subsection{Ablation Studies}

We conduct ablation studies on different components of our Diff-BGM as shown in \cref{tab:obj}. Unconditional means that we use the baseline diffusion model to generate music for each given video. For we do not add any conditions or restrictions, the generation results have the highest diversity(6.421). However, for the lack of control signal and temporal alignment, the quality and correspondence are not so good.
Then we add video feature and feature selector(row 5-6) to the base model. When adding more signals to control the generation process, the quality of the generated music keeps improving and the diversity keeps decreasing. The metric of PCHE has a gain of 0.468, indicating a more clear melody. Besides, with the introduction of video feature, the video-music correspondence score P@20 has a gain of $11.27$, indicating that the music contains information from the video.
In the last row, segment-aware cross-attention layers are added to the model, which focuses on the alignment between music and video and improves the retrieval score. However, when we force the music to pay attention to only short-term context of the video, the music quality decreases.
The results indicate that the quality of music and its correlation with video mutually influence each other when we aim to exert control over the music.

Besides, we conduct an ablation study on the feature selector as shown in \cref{tab:ablation}. In the first two rows, we only use features from one modality (either video dynamic or semantics) to control the generation process. 
We find that when only using language features as condition, the results gain the highest GPS marks (0.641), which means that the structure of generated music is closest to real one and captions facilitate the generation of musical structures.
And in the last two rows, we attempt to use different feature orders to control the generation at different stages. The results indicate that early-stage usage of video semantic features followed by dynamic features yields the best music quality, aligning with the viewpoint that the model generates melody first and then produces rhythm. More ablation studies about the feature selector and SAC Attention can be found in supp.M.

\section{Conclusion}

In this paper, we propose the Diff-BGM framework to tackle the video background music generation task and new evaluation metrics to measure video-music correspondence and also music diversity. We also provide a high-quality dataset, BGM909, comprising temporally and semantically aligned video-music pairs and fine-grained annotations for shots and natural language captions of videos. We address the issue of poor interpretability in existing generative models by using different features to control various stages of the music generation process. We introduce segment-aware cross-attention to temporally align music and video and generate music corresponding to video content and rhythm. Experiments verify that Diff-BGM has the capability of generating high-quality background music for videos.

\raggedright
\noindent\textbf{Acknowledgements.} This work was supported by the grants from the National Natural Science Foundation of China 62372014.

{
    \small
    \bibliographystyle{ieeenat_fullname}
    \bibliography{main}
}

\end{document}